\pdfoutput=1
\documentclass[11pt]{article}
\pdfoutput=1
\usepackage[]{emnlp2021}

\usepackage{times}
\usepackage{latexsym}
\usepackage{graphicx}
\usepackage[T1]{fontenc}
\usepackage[utf8]{inputenc}

\usepackage{microtype}

\newcolumntype{?}{!{\vrule width 1.5pt}}
\usepackage{xurl}
\usepackage{tabularx}
\usepackage{graphicx}
\usepackage{subfig}
\newcommand\cincludegraphics[2][]{\raisebox{-0.3\height}{\includegraphics[#1]{#2}}}
\usepackage{color, colortbl}
\definecolor{LGray}{gray}{0.9}
\definecolor{Gray}{gray}{0.8}
\definecolor{DGray}{gray}{0.7}
\usepackage{xcolor}
\usepackage{booktabs}
\usepackage{amsmath}
\usepackage{amssymb}
\usepackage{xparse}
\NewDocumentCommand \plane{}{\includegraphics[scale=0.1]{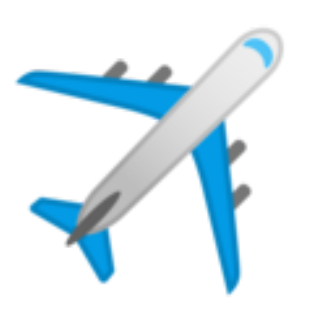}}
\title{Point-of-Interest Type Prediction using Text and Images}

\author{{\bf Danae S\'{a}nchez Villegas} \quad {\bf Nikolaos Aletras}\\
    Computer Science Department, University of Sheffield, UK\\
    {\small
    {\tt \{dsanchezvillegas1, n.aletras\}@sheffield.ac.uk}}\\
    }

\begin{document}
\maketitle
% \begin{abstract}
% The linguistic and multimodal content of a post, such as text and pictures, provide information about the feelings and experiences evoked by participating in an activity or living an experience in a Point-of-Interest (POI), such as a restaurant, park, or airport, which can provide a view of the place's identity. We present a study of multimodal POI type prediction using text and visual information available at the time of posting; predictive models using text, and visual data which significantly outperform the state-of-the-art method for POI type prediction; and an in-depth analysis to uncover POI type characteristics and understand the limitations of our predictive models. The use of linguistic and visual information to infer the type of a POI from social media posts is useful for studies in sociolinguistics, geosemiotics, and cultural geography, and has applications in geosocial networking technologies such as recommendation and visualization systems.
% \end{abstract}

\begin{abstract}
 Point-of-interest (POI) type prediction is the task of inferring the type of a place from where a social media post was shared. Inferring a POI's type is useful for studies in computational social science including sociolinguistics, geosemiotics, and cultural geography, and has applications in geosocial networking technologies such as recommendation and visualization systems. Prior efforts in POI type prediction focus solely on text, without taking visual information into account. However in reality, the variety of modalities, as well as their semiotic relationships with one another, shape communication and interactions in social media. This paper presents a study on POI type prediction using multimodal information from text and images available at posting time. For that purpose, we enrich a currently available data set for POI type prediction with the images that accompany the text messages. Our proposed method extracts relevant information from each modality to effectively capture interactions between text and image achieving a macro F1 of 47.21 across eight categories significantly outperforming the state-of-the-art method for POI type prediction based on text-only methods. Finally, we provide a detailed analysis to shed light on cross-modal interactions and the limitations of our best performing model.\footnote{Code and data are available here: \url{https://github.com/danaesavi/poi-type-prediction}}
\end{abstract}

\section{Introduction}

%The study of the identity of a place concerns research in social and cultural geography \cite{Tuan1991, Benwell2006}, and geographic and visual semiotics by studying \textit{discourse in time and place} (i.e. language as a social practice in terms of linguistic content and the physical, social, and cultural context), as well as linguistic and multimodal representations of a place \cite{kress1996reading, Scollon2003, al20148}.

A place is typically described as a physical space infused with human meaning and experiences that facilitate communication \cite{tuan1977space}. The multimodal content of social media posts (e.g. text, images, emojis) generated by users from specific places such as restaurants, shops, and parks, contribute to shaping a place's identity, by offering information about feelings elicited by participating in an activity or living an experience in that place \cite{Tanasescu2013}.

%%%%%%%%%%%%%%%%%%%%% Sample Tweets %%%%%%%%%%%%%%%%%%%%%%%%%%%%%%

\renewcommand*{\arraystretch}{1.2}
\begin{figure}[!t]
    \footnotesize
    \centering
    %\small
    \begin{tabular}{cc}
        \cincludegraphics[scale=0.055]{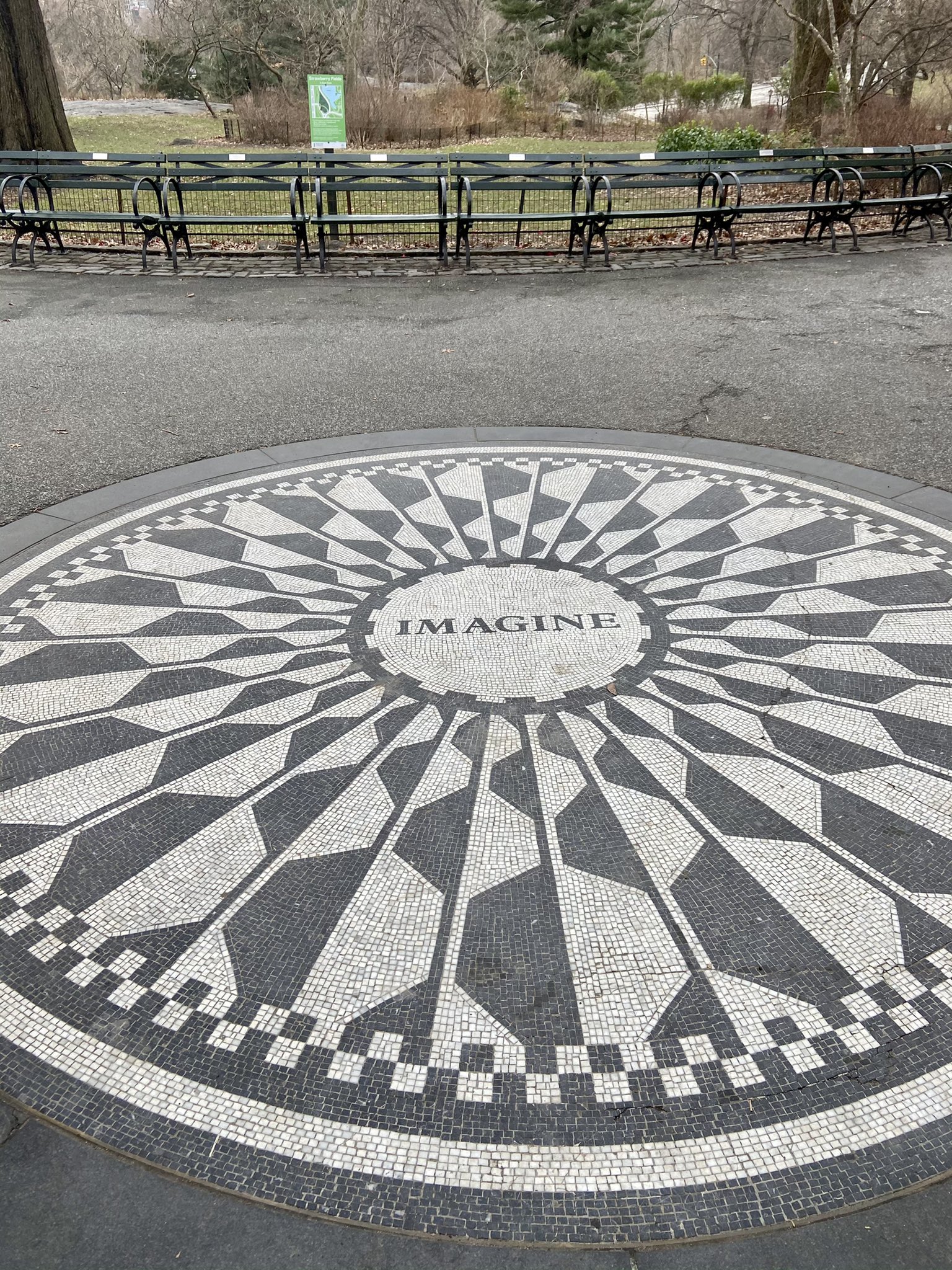}&\cincludegraphics[scale=0.12]{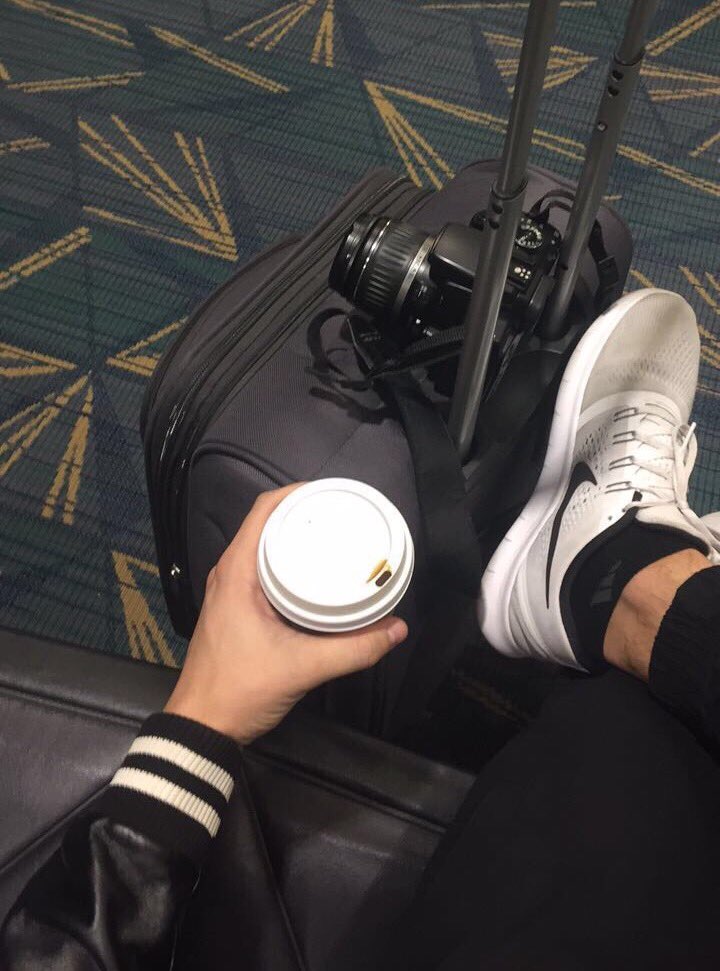}\\
        \begin{tabular}[c]{@{}l@{}}imagine all the people\\ sharing all the world $\sim$ \end{tabular}  & Next stop: NYC \plane{}
    \end{tabular}

    \caption{Example of text and image content of  sample tweets. Users share content that is relevant to their experiences and feelings in the location. }
    \label{fig:samplet}
\end{figure}
%%%%%%%%%%%%%%%%%%%%%%%%%%%%% END Sample Tweets %%%%%%%%%%%%%%%%%%

Fig. \ref{fig:samplet} shows examples of Twitter posts consisting of image-text pairs, shared from two different places or Point-of-Interests (POIs). Users share content that is relevant to their experience in the location. For example, the text \textit{imagine all the people sharing all the world} which is accompanied by a photograph of the Imagine Mosaic in Central Park; and the text \textit{Next stop: NYC} along with a picture of descriptive items that people carry at an airport such as luggage, a camera and a takeaway coffee cup.
%, resulting in meaningful representations of a place's identity.

Developing computational methods to infer the type of a POI from social media posts~\citep{liu2012location,sanchez-villegas-etal-2020-point} is useful for complementing studies in computational social science including sociolinguistics, geosemiotics, and cultural geography \cite{kress1996reading,Scollon2003,al20148}, and has applications in geosocial networking technologies such as recommendation and visualization systems \cite{alazzawi2012can,zhang2018exploiting,van2019go,liu2020exploiting}.

Previous work in natural language processing (NLP) has investigated the language that people use in social media from different locations, by inferring the type of a POI of a given social media post using only text and posting time, ignoring the visual context \cite{sanchez-villegas-etal-2020-point}. %However, there is no previous study that includes visual context.
However, communication and interactions in social media are naturally shaped by the variety of available modalities and their semiotic relationships (i.e. how meaning is created and communicated) with one another \cite{georgakopoulou2015routledge, kruk-etal-2019-integrating, vempala-preotiuc-pietro-2019-categorizing}.

% The use of linguistic and visual information to infer the type of a POI from social media posts is useful for studies in sociolinguistics, geosemiotics, and cultural geography \cite{kress1996reading,Scollon2003,al20148}, and has applications in geosocial networking technologies such as recommendation and visualization systems \cite{alazzawi2012can,zhang2018exploiting,van2019go,liu2020exploiting}.

 In this paper, we propose POI type prediction using multimodal content available at posting time by taking into account textual and visual information. Our contributions are as follows:

 \begin{itemize}
     \item We enrich a publicly available data set of social media posts and POI types with images;

     \item We propose a multimodal model that combines text and images in two levels using: (i) a modality gate to control the amount of information needed from the text and image; (ii) a cross-attention mechanism to learn cross-modal interactions. Our model significantly outperforms the best state-of-the-art method proposed by \citet{sanchez-villegas-etal-2020-point};

     \item We provide an in-depth analysis to uncover the limitations of our model and uncover cross-modal characteristics of POI types.
 \end{itemize}

 %predictive models using text, and visual data which significantly outperform the state-of-the-art method for POI type prediction; and an in-depth analysis to uncover POI type characteristics and understand the limitations of our predictive models. %Previous text-image classification in social media requires that the data is fully paired, i.e. every post contains an image and a text \cite{nguyen-shirai-2015-topic, chambers-etal-2015-identifying, cai-etal-2019-multi, wang-etal-2020-cross-media}. However, this requirement may not be satisfied since not all posts contain both modalities \footnote{\url{https://buffer.com/resources/twitter-data-1-million-tweets/}}. This work considers both cases, (1) all modalities (text-image pairs) are available, and content in only one modality (text) is available.

 \section{Related Work}

% \subsection{Identity Studies in Social Media}
% % identity: Not sure if we need this section or maybe use it in the intro
% Previous studies of \textit{identity} in social media are focused on studying how users choose to portray themselves to others on a social media platform \cite{priante-etal-2016-whoami,georgalou2017discourse,li-etal-2020-emoji}. \citet{georgalou2017discourse} investigates how identity is discursively constructed within Facebook and how representations of identity are manifested. \citet{li-etal-2020-emoji} presents a study of the relationship between visual characters (emojis) included in Twitter bios and the user’s online self-identity. Rather than a user-based study, our research aims to uncover the characteristics associated with various types of POIs.

\subsection{POI Analysis}
%poi types
POIs have been studied to classify functional regions (e.g. residential, business, and transportation areas) and to analyze activity patterns using social media check-in data and geo-referenced images \cite{zhi2016latent,liu2020visualizing,zhou2020poi,zhang2020uncovering}. \citet{zhou2020poi} presents a model for classifying POI function types (e.g. bank, entertainment, culture) using POI names and a list of results produced by searching for the POI name in a web search engine. \citet{zhang2020uncovering} makes use of social media check-ins and street-level images to compare the different activity patterns of visitors and locals, and uncover inconspicuous but interesting places for them in a city. A framework for extracting emotions (e.g. joy, happiness) from photos taken at various locations in social media is described in \citet{kang2019extracting}. %they demonstrate its utility for understanding the sense of place in geography.

\subsection{POI Type Prediction}

POI type prediction is related to geolocation prediction of social media posts that has been widely studied in NLP \cite{eisenstein-etal-2010-latent, roller-etal-2012-supervised, dredze-etal-2016-geolocation}. However, while geolocation prediction aims to infer the exact geographical location of a post using language variation and geographical cues, POI type prediction is focused on identifying the characteristics associated with each type of place, regardless of its geographic location.

Previous work on POI type prediction from social media content has used Twitter posts (text and posting time), to identify the POI type from where a post was sent from \cite{liu2012location,sanchez-villegas-etal-2020-point}. \citet{liu2012location} incorporate text, temporal features (posting hour) and user history information into probabilistic text classification models. Rather than a user-based study, our research aims to uncover the characteristics associated with various types of POIs. \citet{sanchez-villegas-etal-2020-point} analyze semantic place information of different types of POIs by using text and temporal information (hour, and day of the week) of a Twitter's post. To the best of our knowledge, this is the first study to combine textual and visual features to classify POI types (e.g. arts \& entertainment, nightlife spot) from social media messages, regardless of its geographic location.

%%%%%%%%%%%%%%%%%%%%%%%%%%%%%%% Data TABLE %%%%%%%%%%%%%%%%%%%%%%%%%
% TODO: add percentages of images and general header for each split
\begin{table*}[t!]
\resizebox{\linewidth}{!}{
\centering
\small
\begin{tabular}{ l |r r| r r| r r| r }
\hline
\rowcolor[HTML]{C0C0C0}
  & \multicolumn{2}{c}{\textbf{Train}} & \multicolumn{2}{c}{\textbf{Dev}} &  \multicolumn{2}{c}{\textbf{Test}}  &   \\
\rowcolor[HTML]{C0C0C0}
\textbf{Category}  &\textbf{\# Tweets} &\textbf{ \# Images} & \textbf{\# Tweets} &\textbf{ \# Images} & \textbf{\# Tweets} & \textbf{\# Images} & \textbf{Tokens}  \\ \midrule
\textbf{Arts \& Entertainment}        &40,417 & 20,711 & 4,755 &2,527 & 5,284 &2,740 & 14.41      \\ \hline
\rowcolor[HTML]{EFEFEF}
\textbf{College \& University}        &21,275 &9,112 & 2,418 &1,057  & 2,884 &1,252  & 15.52     \\ \hline
\textbf{Food}                         &6,676 &2,969 & 869 &351  & 724 &280    & 14.34      \\ \hline
\rowcolor[HTML]{EFEFEF}
\textbf{Great Outdoors}               &27,763 &13,422 & 4,173 &2,102  & 3,653 & 1,948 & 13.49 \\ \hline
\textbf{Nightlife Spot}               &5,545 &2,532 & 876 &385  & 656 &353 & 15.46 \\ \hline
\rowcolor[HTML]{EFEFEF}
\textbf{Professional \& Other Places} &30,640 &13,888 & 3,381 &1,499  & 3,762 &1,712  & 16.46 \\ \hline
\textbf{Shop \& Service}              &8,285 &3,455 & 886 &266  & 812 &353 & 15.31    \\ \hline
\rowcolor[HTML]{EFEFEF}
\textbf{Travel \& Transport}          &16,428 &6,681 & 2,201 &829  & 1,872 &789  & 14.88       \\
\textbf{All}          &157,029 &72,679 (46.28\%) & 19,559  &9,006 (46.05\%)  & 19,647&9,410 (47.90\%) & 14.92
\\\hline
\end{tabular}}

\caption{POI categories and data set statistics showing the number of tweets for each category, and number (\%) of tweets having an accompanying image}
\label{tab:stats}
\end{table*}
%%%%%%%%%%%%%%%%%%%%%%%%%%%%%%% END Data TABLE %%%%%%%%%%%%%%%%%%%%%%

\subsection{Social Media Analysis using Text and Images}
%discuss meaning multiplication \cite{bateman2014text}, emphasizing that “the text+image integration requires inference that creates a new meaning
The combination of text and images of social media posts has been largely used for different applications such as sentiment analysis, \cite{nguyen-shirai-2015-topic, chambers-etal-2015-identifying}, sarcasm detection \cite{cai-etal-2019-multi} and text-image relation classification \cite{vempala-preotiuc-pietro-2019-categorizing, kruk-etal-2019-integrating}. \citet{moon-etal-2018-multimodal-named} propose a model for recognizing named entities from short social media texts using image and text. \citet{cai-etal-2019-multi} use a hierarchical fusion model to integrate image and text context with an attention-based fusion. \citet{chinnappa-etal-2019-extracting} examine the possession relationships from text-image pairs in social media posts. % As an additional textual input, they extract objects and events identified in an image.
\citet{wang-etal-2020-cross-media} use texts and images for predicting the keyphrases (i.e. representative terms) for a post by aligning and capturing the cross-modal interactions via cross-attention. Previous text-image classification in social media requires that the data is fully paired, i.e. every post contains an image and a text. However, this requirement may not be satisfied since not all posts contain both modalities \footnote{\url{https://buffer.com/resources/twitter-data-1-million-tweets/}}. This work considers both cases, (1) all modalities (text-image pairs) are available, and content in only one modality (text or image) is available.

Social media analysis research has also looked at the semiotic properties of text-image pairs in posts~\cite{alikhani-etal-2019-cite,vempala-preotiuc-pietro-2019-categorizing,kruk-etal-2019-integrating}. \citet{vempala-preotiuc-pietro-2019-categorizing} investigate the relationship between text and image content by identifying overlapping meaning in both modalities, those where one modality contributes with additional details, and cases where each modality contributes with different information. \citet{kruk-etal-2019-integrating} analyze the relationship between the text-image pairs and find that when the image and caption diverge semiotically, the benefit from multimodal modeling is greater.

\section{Task \& Data}
\label{sec:task}
%\section{Task}
\citet{sanchez-villegas-etal-2020-point} define POI type prediction as a multi-class classification task where given the text content of a post, the goal is to classify it in one of the $M$ POI categories. In this work, we extend this task definition to include images in order to capture the semiotic relationships between the two modalities. For that purpose, we consider a social media post $P$ (e.g. tweet) to comprise of a text and image pair ${(x^t,x^v)}$, where $x^t \in \mathbb{R}^{d_t}$ and $x^v \in \mathbb{R}^{d_v}$ are the textual and visual vector representations respectively.
%The aim is to predict the POI category $m \in M$ given $P$.

\subsection{POI Data}
We use the data set introduced by \citet{sanchez-villegas-etal-2020-point} which contains $196,235$ tweets written in English, labeled with one out of the eight POI broad type categories shown in Table \ref{tab:stats}, which correspond to the 8 primary top-level POI categories in `Places by Foursquare', a database of over
105 million POIs worldwide managed by Foursquare. To generalize to locations not present in the training set, we use the same location-level data splits (train, dev, test) as in \citet{sanchez-villegas-etal-2020-point}, where each split contains tweets from different locations.

\subsection{Image Collection} We use the Twitter API to collect the images that accompany each textual post in the data set. For the tweets that have more than one image, we select the first available only. This results in $91,224$ tweets with at least one image. During the image processing (see Section \ref{sec:img}) we removed 129 images because we found they were either damaged, absent\footnote{Removed by Twitter due to violations to the Twitter Rules and Terms of Service.}, or no objects were detected, resulting in $91,095$ text-image pairs (see Table \ref{tab:stats} for data statistics). In order to deal with the rest of the tweets with no associated image, we pair them with a single `average' image computed over all images in the train set: $x^v = avg(x^v_{tr})$. The intuition behind this approach is to generate a `noisy' image that is not related and does not add to the meaning \cite{vempala-preotiuc-pietro-2019-categorizing}.\footnote{Early experimentation with associating tweets with the image of the most similar tweet that contains a real image from the training data yielded similar performance.} %This is one of the relationships in this paper
%In the following sections, we will refer to the subset of tweets that are originally accompanied by an image as \emph{Text-Image Only Data}, and the complete POI data collection as \emph{All POI Data}.

\subsection{Exploratory Analysis of Image Data} To shed light on the characteristics of the collected  images, we apply object detection on the images collected using Faster-RCNN \cite{ren2016faster} pretrained on Visual Genome \cite{krishna2017visual, anderson2018bottom}. Table \ref{tab:objects} shows the most common objects for each specific category. We observe that most objects are related to items one would find in each place category (e.g. `spoon', `meat', `knife' in \emph{Food}). Clothing items are common across category types (e.g. `shirt', `jacket', `pants') suggesting the presence of people in the images. A common object tag of the \emph{Shop \& Service} category is `letters', which concerns images that contain embedded text. Finally, the category \emph{Great Outdoors} includes object tags such as `cloud', `hill', and `grass', words that describe the landscape of this type of place.

%%%%%%%%%%%%%%%%%%%%%%%%%%%%%%% Objects TABLE %%%%%%%%%%%%%%%%%%%%%%%%%%%%
\begin{table}[t!]
\resizebox{\linewidth}{!}{
\centering
\begin{tabular}{ l l }
\hline
\rowcolor[HTML]{C0C0C0}
\textbf{Category}  &\textbf{Common Objects in Images}   \\ \hline
\textbf{Arts \& Entertainment}        &\begin{tabular}[c]{@{}l@{}} light, pants,  shirt, arm, picture,\\  hair, glasses, line, girl, jacket \end{tabular}   \\ \hline
\rowcolor[HTML]{EFEFEF}
\textbf{College \& University}        &\begin{tabular}[c]{@{}l@{}} pants, shirt,  line, hair, arm,\\ picture,  light,  glasses, girl, trees \end{tabular} \\ \hline
\textbf{Food}                         & \begin{tabular}[c]{@{}l@{}} cup, picture, spoon, meat, knife, \\  arm, glasses, shirt, pants, handle \end{tabular} \\ \hline
\rowcolor[HTML]{EFEFEF}
\textbf{Great Outdoors}               & \begin{tabular}[c]{@{}l@{}} trees, arm, pants, cloud, hill,\\ line, shirt, grass, picture, glasses \end{tabular}\\ \hline
\textbf{Nightlife Spot}               &\begin{tabular}[c]{@{}l@{}} arm, picture, shirt, light, hair,\\ pants, glasses, mouth, girl, cup \end{tabular}  \\ \hline
\rowcolor[HTML]{EFEFEF}
\textbf{Professional \& Other Places} &\begin{tabular}[c]{@{}l@{}} pants, shirt, picture, light, hair,\\ screen, line, arm, glasses, girl \end{tabular}  \\ \hline
\textbf{Shop \& Service}              &\begin{tabular}[c]{@{}l@{}} picture, pants,  arm, shirt, glasses,\\ light, hair, line, girl, letters \end{tabular} \\ \hline
\rowcolor[HTML]{EFEFEF}
\textbf{Travel \& Transport}          &\begin{tabular}[c]{@{}l@{}} pants, shirt, light, screen, arm,\\ hair, glasses, picture, chair, line \end{tabular}
%\textbf{All}          &\begin{tabular}[c]{@{}l@{}} person ,  wall ,  building ,  man ,  ground ,\\  letter ,  ceiling ,  window ,  logo ,  shoe  \end{tabular}
\\\hline
\end{tabular}}

\caption{Most common objects for each POI category.}
\label{tab:objects}
\end{table}
%%%%%%%%%%%%%%%%%%%%%%%%%%%%%%% END Objects TABLE %%%%%%%%%%%%%%%%%%%%%%

%%%%%%%%%%%%%%%%%%% Methodology %%%%%%%%%%%%%%%%%%%%%%%%%%%%%%%
\section{Multimodal POI Type Prediction}
\label{sec:mmpoi}
%Let $x_t \in \mathbb{R}^{d_t}$ $x_v \in \mathbb{R}^{d_v}$.
\subsection{Text and Image Representation}
Given a text-image post $P = (x^t,x^v)$, $x^t \in \mathbb{R}^{d_t}$, $x^v \in \mathbb{R}^{d_v}$, we first compute text and image encoding vectors $f^t$, $f^v$ respectively.

\paragraph{Text} We use Bidirectional Encoder Representations from Transformers (BERT) \cite{devlin-etal-2019-bert} to obtain the text feature representations $f^t$ by extracting the `classification' [CLS] token.

\paragraph{Image} For encoding the images, we use Xception \cite{chollet2017xception} pre-trained on ImageNet \cite{5206848}.\footnote{Early experimentation with ResNet101 \cite{he2016deep} and EfficientNet \cite{pmlr-v97-tan19a} yielded similar results.} We extract convolutional feature maps for each image and we apply average pooling to obtain the image representation $f^v$.

\begin{figure*}[t!]
    \centering
    \includegraphics[width=12cm]{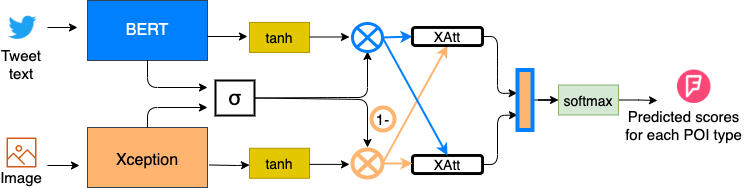}
    \caption{Overview of our MM-Gated-XAtt model which combines features from text and image modalities for POI type prediction.}
    \label{fig:model}
\end{figure*}

\subsection{MM-Gate}
% to manage the information flow from each modality,
Given the complex semiotic relationship between text and image, we need a
weighting strategy that assigns more importance to the most relevant modality while suppressing irrelevant information. Thus, a first approach is to use gated multimodal fusion (MM-Gate), similar to the approach proposed by \citet{arevalo2020gated} to control the contribution of text and image to the POI type prediction. Given $f^t$, $f^v$ the text and visual vectors, we obtain the multimodal representation $h$ of a post $P$ as follows:
\begin{align}
    h^t=tanh(W^tf^t+b^t)  \label{eq:gate1} \\
    h^v=tanh(W^vf^v+b^v)  \label{eq:gate2} \\
    z=\sigma(W^z[f^t;f^v]+b^z) \label{eq:gate3} \\
    h=z*h^t+(1-z)*h^v \label{eq:gate4}
\end{align}

\noindent where $W^t \in \mathbb{R}^{d_t}$, $W^v \in \mathbb{R}^{d_v}$ and $W^z \in R^{d_t+d_v}$ are learnable parameters, \emph{tanh} is the activation function and $h^t,h^v \in \mathbb{R^d}$ are projections of $f^t$ and $f^v$. $[;]$ denotes concatenation and $\sigma$ is the sigmoid activation function. $h$ is a weighted combination of the textual and visual information $h^t$ and $h^v$ respectively. We fine-tune the entire model by adding a classification layer with a softmax activation function for POI type prediction

\subsection{MM-XAtt}

The MM-Gate model does not capture interactions between text and image that might be beneficial for learning semiotic relationships. To model cross-modal interactions, we adapt the cross-attention mechanism \cite{tsai-etal-2019-multimodal, tan-bansal-2019-lxmert} to combine text and image information for multimodal POI type prediction (MM-XAtt). Cross-attention consists of two attention layers, one from textual $f^t$ to visual features $f^v$ and one from visual to textual features. We first linearly project the text and visual representations to obtain the same dimensionality ($d_{proj}$). Then, we compute the scaled dot attention ($a=softmax\frac{(Q(K)^T)}{\sqrt{d_{proj}}}V$) with the projected textual vector as query ($Q$), and the projected image vector as the key ($K$) and values ($V$), and vice versa. The multimodal representation $h$ is the sum of the resulting attention layers. The entire model is fine-tuned by adding a classification layer with a softmax activation function.

\subsection{MM-Gated-XAtt}
%\subsection{BERT-Img$_{\text{Gate-CrossAtt}}$}
%\subsection{Predictive Models}
\citet{vempala-preotiuc-pietro-2019-categorizing} have demonstrated that the relationship between the text and image in a social media post is complex. Images may or may not add meaning to the post and the text content (or meaning) may or may not correspond to the image. We hypothesize that this might actually happen in posts made from particular locations, i.e. language and visual information may or may not be related. To address this, we propose (1) using gated multimodal fusion to manage the flow of information from each modality, and (2) also learn cross-modal interactions by using cross-attention on top of the gated multimodal mechanism. Fig. \ref{fig:model} shows an overview of our model architecture (MM-Gated-XAtt). Given the text and image representations $f^t$, $f^v$ respectively, we compute $h^t$, $h^v$, and $z$ as in Equation \ref{eq:gate1}, \ref{eq:gate2} and \ref{eq:gate3}. Next, we apply cross-attention using two attention layers where the query and context vectors are the weighted representations of the text and visual modalities, $z*h^t$ and $(1-z)*h^v$, and vice versa. The multimodal context vector $h$ is the sum of the resulting attention layers. Finally, we fine-tune the model by passing $h$ through a classification layer for POI type prediction with a softmax activation function.

%%%%%%%%%%%%%%%%%%%%  All Tweets (Avg) %%%%%%%%%%%%%%%%%%%%%%%%%%%%%%%%%%%%%
\begin{table*}[!t]
\centering
\tiny
\resizebox{0.80\textwidth}{!}{
\begin{tabular}{|l|l|l|l|}
%\begin{tabular}{|l?l|l|l?l|l|l|}
\hline
\rowcolor[HTML]{C0C0C0}
%\multicolumn{1}{|c|}{\cellcolor[HTML]{C0C0C0}} &
%\multicolumn{3}{|c|}{\cellcolor[HTML]{C0C0C0}\textbf{All POI Data}} & \multicolumn{3}{|c|}{\cellcolor[HTML]{C0C0C0}\textbf{Text-Image Only}}\\ \hline
\multicolumn{1}{|c|}{\cellcolor[HTML]{C0C0C0}\textbf{Model}} &
\multicolumn{1}{|c|}{\cellcolor[HTML]{C0C0C0}\textbf{F1}} & \multicolumn{1}{|c|}{\cellcolor[HTML]{C0C0C0}\textbf{P}} & \multicolumn{1}{|c|}{\cellcolor[HTML]{C0C0C0}\textbf{R}} %&
%\multicolumn{1}{|c|}{\cellcolor[HTML]{C0C0C0}\textbf{F1}} & \multicolumn{1}{|c|}{\cellcolor[HTML]{C0C0C0}\textbf{P}} & \multicolumn{1}{|c|}{\cellcolor[HTML]{C0C0C0}\textbf{R}}
\\ \hline

%\multicolumn{7}{|c|}{\cellcolor[HTML]{C0C0C0}\textbf{Baselines}} \\ \hline
Majority                          & 5.30      & 3.36      & 12.50  %& 5.62      & 3.63      & 12.50
\\
\hline
%\citet{sanchez-villegas-etal-2020-point}&&&&&&\\
%LR-T                             & 14.01    & 15.78     & 16.06 \\
BERT \citep{sanchez-villegas-etal-2020-point}                            & 43.67 (0.01)    & \textbf{48.44} (0.02)     & 41.33 (0.01) %& 39.21 (x.xx)    & 38.29 (x.xx)     & 44.16 (x.xx)
\\
%BERT-TS                          & 43.47 (0.007)    & 48.40 (0.006)    & 41.26 (0.008)
\hline
% Time$_{emb}$                     & 15.60    & 16.49     & 16.45 \\
% BERT-Time$_{emb}$                & 40.47    & 40.27     & 44.76 \\
% BERT-Time$_{emb-GLU}$            & 42.11    & 41.59     & 45.94 \\ \hline

ResNet                           & 21.11 (1.81)     & 23.23 (2.09)     & 29.90 (3.31) %& 34.02 (2.99)     & 40.02 (3.37)    & 35.89 (1.78)
\\
EfficientNet                     & 24.72 (0.76)     & 28.05 (0.28)    & 35.48 (0.23)  %& 37.74 (0.07)     & 44.76 (0.42)     & 38.57 (0.18)
\\
Xception                         & 23.64 (0.44)     & 25.62 (0.50)     & 34.12 (0.49) %& 36.17 (0.77)     & 40.03 (0.55)     & 35.80 (0.44)
\\
%FasterRCNN                       & 23.65 (0.53)     & 23.99 (0.01)     & 32.81 (0.28) \\
\hline

Concat-BERT+ResNet           & 43.28 (0.37)     & 42.72 (0.51)     & 47.59 (0.45)  %& 44.77 (2.31)     & 43.92 (1.56)     & 49.15 (1.56
\\
Concat-BERT+EfficientNet          & 41.56 (0.71)     & 41.54 (0.88)     & 43.97 (0.79) %& 45.69 (0.67)     & 44.51 (0.40)     & 50.49 (1.19)
\\
Concat-BERT+Xception               & 44.00 (0.52)     & 43.34 (0.70)     & 48.35 (0.75)  %& 45.15 (0.91)     & 44.56 (0.39)     & 48.02 (1.47)
\\
Attention-BERT+Xception                  & 42.89 (0.44)    &  42.74 (0.19)    & 46.78 (1.28) %&&&
\\
%BERT+ResNet$_{CrossAtt}$            &  17.86    &  15.60    & 25.10 \\
%BERT+EffNet$_{CrossAtt}$            &  15.52    &  26.82    & 22.36 \\
Guided Attention-BERT+Xception         &  41.53 (0.57)    & 41.10 (0.55)    & 45.36 (0.48)  %& 44.07 (0.63)   & 42.93 (0.69)      & 48.59 (0.41)
\\ \hline

LXMERT                           & 40.17 (0.62)     & 40.26 (0.24)     & 42.25 (2.38) %& 47.72 (0.98)   & 47.00 (0.15)     & 50.71 (1.27)
\\
Ensemble-BERT+LXMERT        & 43.82 (0.47)    & 43.50 (0.20)     & 44.67 (0.66) %&-&-&-
\\
\bottomrule
MM-Gate                  & 44.64 (0.65)     & 43.67 (0.49)     & 48.50 (0.18) %& 45.87 (1.48)     & 45.11 (1.08)     & 48.68 (0.63)
\\
MM-XAtt                &  27.31 (1.58)    &  37.06 (2.66)    & 29.71 (0.60) %& 48.93 (2.08)   & 48.23 (2.71)     & 56.50 (1.89)
\\
MM-Gated-XAtt (Ours)            & \textbf{47.21$\dagger$} (1.70)     & 46.83 (1.45)     & \textbf{50.69} (2.21) %& \textbf{57.64} (3.64)     %& \textbf{56.53} (3.58  & \textbf{62.06} (2.25)
\\ %\hline

%Ensemble-BERT+MM-Gated-XAtt        & 48.54 (1.72)    & 48.07 (2.05)     & 49.84 (1.08) %&-&-&-\\
\hline
% BERT+Xc-T$_{CONCAT}$             & \underline{45.02}     & 44.34     & 47.62\\
% BERT+Xc-T$_{GLU}$                & \underline{44.14}     & 43.45     & 48.33 \\ \hline

\end{tabular}}
\caption{Macro F1-Score, precision (P) and recall (R) for POI type prediction ($\pm$ std. dev.) Best results are in bold. $\dagger$ indicates statistically significant improvement (t-test, $p<0.05$) over BERT \cite{sanchez-villegas-etal-2020-point}.}
\label{tab:resultsfull}
\end{table*}
%%%%%%%%%%%%%%%%%%%% End All Results %%%%%%%%%%%%%%%%%%%%%%%%%%%%%%%%%

%%%%%%%%%%%%%%%%%%% Experimental Setup %%%%%%%%%%%%%%%%%%%%%%%%%%%%%%%

\section{Experimental Setup}
\label{sec:methods}

\subsection{Baselines}
%This section describes different baseline models we apply for POI type prediction. We first experiment with text-only and image-only methods, and then we explore text-image fusion models including competitive multimodal baselines such as LXMERT \cite{tan-bansal-2019-lxmert}.

We compare our models against (1) text-only; (2) image-only; and (3) other state-of-the-art multimodal approaches.\footnote{We include a majority class baseline (i.e. assigning all instances in the test set the most frequent label in the train set).}

\paragraph{Text-only} We fine-tune BERT for POI type classification by adding a classification layer with softmax activation function on top of the [CLS] token which is the best performing model in \citet{sanchez-villegas-etal-2020-point}.

\paragraph{Image-only} We fine-tune three pre-trained models that are popular in various computer vision classification tasks: (1) ResNet101 \cite{he2016deep}; (2) EfficientNet \cite{pmlr-v97-tan19a}; and (3) Xception \cite{chollet2017xception}. Each model is fine-tuned on POI type classification by adding an output softmax layer.
%addresses the vanishing gradient problem that occurs when training deep neural networks, by introducing the \textit{identity shortcut connection}, which skips one or more layers.

\paragraph{Text and Image}
%We explore different fusion strategies: concatenation (BERT-Img$_{\text{Concat}}$), attention mechanism (BERT-Img$_{\text{Att}}$), and guided attention (BERT-Img$_{\text{GuidedAtt}}$).
For combining text and image information, we experiment with different standard fusion strategies: (1) we project the image representation $f^v$, to the same dimensionality as $f^t \in \mathbb{R}^{d_t}$ using a linear layer and then we concatenate the vectors (\textbf{Concat}); (2) we project the textual and visual features to the same space and then we apply self-attention to learn weights for each modality (\textbf{Attention}); (3) we also adapt the guided attention introduced by \citet{anderson2018bottom} for learning attention weights at the object-level (and other salient regions) rather than equally sized grid-regions (\textbf{Guided Attention}); (4) we compare against  \textbf{LXMERT}, a transformer-based model that has been pre-trained on text and image pairs for learning cross-modality interactions~\citep{tan-bansal-2019-lxmert}. All models are fine-tuned by adding a classification layer with a softmax activation function for POI type prediction. Finally, we evaluate a simple ensemble strategy by using LXMERT for classifying tweets that are originally accompanied by an image and BERT for classifying text-only tweets (\textbf{Ensemble}).

\subsection{Text Processing}
We use the same tokenization settings as in \citet{sanchez-villegas-etal-2020-point}. For each tweet, we lowercase text and replace URLs and @-mentions of users with placeholder tokens.

\subsection{Image Processing}
\label{sec:img}
Each image is resized to ($224\times224$) pixels representing a value for the red, green and blue color in the range of $[0,255]$. The pixel values of all images are normalized. For LXMERT and Guided Attention fusion, we extract \textit{object-level} features using Faster-RCNN \cite{ren2016faster} pretrained on Visual Genome \cite{krishna2017visual} following \citet{anderson2018bottom}. We keep 36 objects for each image as in \citet{tan-bansal-2019-lxmert}.%\footnote{During this process we removed 129 images because we found they were either damaged, absent or no objects were detected.}

\subsection{Implementation Details}
We select the hyperparameters for all models using early stopping by monitoring the validation loss using the Adam optimizer~\citep{kingma2014adam}. Because the data is imbalanced, we estimate the class weights using the `balanced' heuristic \cite{king2001logistic}. All experiments are performed using a Nvidia V100 GPU. %For hyperparameter selection, see Appendix~\ref{sec:appendix}.

%\subsection{Hyperparameter details}
%\label{sec:hyperparams}
\paragraph{Text-only} We fine-tune  BERT for 20 epochs and choose the epoch with the lowest validation loss. We use the pre-trained base-uncased model for BERT \cite{Vaswani2017, devlin-etal-2019-bert} from HuggingFace library (12-layer, 768-dimensional) with a maximal sequence length of 50 tokens. We fine-tune BERT for 2 epochs and learning rate $\eta=2e^{-5}$ with $\eta \in \{2e^{-5},3e^{-5},5e^{-5}\}$.

\paragraph{Image-only} For ResNet101, we fine-tune for 5 epochs with learning rate $\eta=1e^{-4}$ and dropout $\delta=0.2$ ($\delta$ in $[0,0.5]$ using random search) before passing the image representation through the classification layer. EfficientNet is fine-tuned for 7 epochs with $\eta=1e^{-5}$ and $\delta=0.5$. Xception is fine-tuned for 6 epochs with $\eta=1e^{-5}$ and $\delta=0.5$.

\paragraph{Text and Image} Concat-BERT+Xception, Concat-BERT+ResNet and Guided Attention-BERT+Xception are fine-tuned for 2 epochs with $\eta=1e^{-5}$ and $\delta=0.25$; Concat-BERT+EfficientNet for 4 epochs with $\eta=1e^{-5}$ and $\delta=0.25$; Attention-BERT+Xception for 3 epochs with $\eta=1e^{-5}$ and $\delta=0.25$; MM-XAtt for 3 epochs with $\eta=1e^{-5}$ and $\delta=0.15$; MM-Gate and MM-Gated-XAtt for 2 epochs with $\eta=1e^{-5}$ and $\delta=0.05$; $\eta \in \{2e^{-5},3e^{-5},5e^{-5}\}$, $\delta$ from $[0,0.5]$ (random search) before passing through the classification layer. The dimensionality of the multimodal representation $h$ (Eq. \ref{eq:gate4}) is set to 200. We fine-tune LXMERT for 4 epochs with $\eta=1e^{-5}$ where $\eta \in \{1e^{-3},1e^{-4},1e^{-5}\}$ and dropout $\delta =0.25$ ($\delta$ in $[0,0.5]$, random search) before passing through the classification layer.

\subsection{Evaluation}
We evaluate the performance of all models using macro F1, precision, and recall. Results are obtained over three runs using different random seeds reporting the average and the standard deviation.

\section{Results}
\label{sec:res}
% TODO: train the models on text image pairs and test with all data
%This section contains the experimental results obtained on \emph{All POI Data} as well as on the subset of tweets that are originally accompanied by an image (\emph{Text-Image Only Data}), in order to investigate the effects of each method in both cases: (1) trained (and tested) on image-text pairs, and (2) a more representative of a real-world scenario in social media where not all tweets are accompanied by an image.

The results of POI type prediction are presented in Table \ref{tab:resultsfull}. We first examine the impact of each modality by analyzing the performance of the unimodal models, then we investigate the effect of multimodal methods for POI type prediction, and finally we examine the performance of our proposed model MM-Gated-XAtt by analyzing each component independently.

We observe that the text-only model (BERT) achieves 43.67 F1 which is substantially higher than the performance of image-only models (e.g. the best performing EfficientNet model obtains 24.72 F1). This suggests that text encapsulates more relevant information for this task than images on their own, similar to other studies in multimodal computational social science \cite{wang-etal-2020-cross-media,ma-etal-2021-effectiveness}.

Models that simply concatenate text and image vectors have close performance to BERT ($44.0$ for Concat-BERT+Xception) or lower ($41.56$ for Concat-BERT+EfficientNet). This suggests that assigning equal importance to text and image information can deteriorate performance. It also shows that modeling cross-modal interactions is necessary to boost performance of POI type classification models.

Surprisingly, we observe that the pre-trained multimodal LXMERT fails to improve over BERT (40.17 F1) while its performance is lower than simpler concatenative fusion models. We speculate that this is because LXMERT is pretrained on data where both, text and image modalities share common semantic relationships which is the case in  standard vision-language tasks including image captioning and visual question answering \cite{Zhou_Palangi_Zhang_Hu_Corso_Gao_2020,NEURIPS2019_c74d97b0}. On the other hand, text-image relationships in social media data for inferring the type of location from which a message was sent are more diverse, highlighting the particular challenges for modeling text and images together \cite{hessel-lee-2020-multimodal}.

% %%%%%%%%%%%%%%%%%%%% With Image Only %%%%%%%%%%%%%%%%%%%%%%%%%%%%%%%%%%%%%
\begin{table}[!t]
\centering
\small
\resizebox{0.38\textwidth}{!}{
\begin{tabular}{|l|l|}
\hline
\rowcolor[HTML]{C0C0C0}
\multicolumn{2}{|c|}{\cellcolor[HTML]{C0C0C0}\textbf{Text-Image Only}} \\ \hline
\multicolumn{1}{|c|}{\cellcolor[HTML]{C0C0C0}\textbf{Model}} &
\multicolumn{1}{c|}{\cellcolor[HTML]{C0C0C0}\textbf{F1}} \\ \hline
LXMERT              & 47.72 (0.98)
\\
MM-Gate             & 45.87 (1.48)
\\
MM-XAtt             & 48.93 (2.08)
\\
MM-Gated-XAtt (Ours)  &   \textbf{57.64} (3.64)   \\ \hline

\end{tabular}}
\caption{Macro F1-Score for POI type prediction on tweets that are originally accompanied by an image. Best results are in bold.}
\label{tab:resultsimgs}
\end{table}
%%%%%%%%%%%%%%%%%%%% End With Image Only %%%%%%%%%%%%%%%%%%%%%%%%%%%%%%%%%

Our proposed MM-Gated-XAtt model achieves $47.21$ F1 which  significantly (t-test, $p<0.05$) improves over BERT, the best performing model in \citet{sanchez-villegas-etal-2020-point} and consistently outperforms all other image-only and multimodal approaches. This confirms our main hypothesis that modeling text with image jointly to learn the interactions between modalities benefit performance in POI type prediction. We also observe that using only the gating mechanism (MM-Gate) outperforms (44.64 F1) all other models except for MM-Gated-XAtt. This highlights the importance of controlling the information flow for the two modalities. Using cross-attention on its own (MM-XAtt), on the other hand, fails to improve over other multimodal approaches, implying that learning cross-modal interactions is not sufficient on its own. This supports our hypothesis that language and visual information in posts sent from specific locations may be or may not be related, and that managing the flow of information from each modality improves the classifier's performance.

 Finally, we investigate using less noisy text-image pairs in alignment with related computational social science studies involving text and images \cite{moon-etal-2018-multimodal-named,cai-etal-2019-multi,chinnappa-etal-2019-extracting}. We train and test LXMERT, MM-Gate, MM-XAtt, and MM-Gated-XAtt on tweets that are originally accompanied by an image (see Section \ref{sec:task}), excluding all text-only tweets. The results are shown in Table \ref{tab:resultsimgs}. In general, performance is higher for all models using less noisy data. Our proposed model MM-Gated-XAtt consistently achieves the best performance ($57.64$ F1). In addition, we observe that LXMERT and MM-XAtt produce similar results ($47.72$ and $48.93$ F1 respectively) suggesting that cross-attention can be applied directly to text-image pairs in low-noise settings without hurting the model performance. The benefit of controlling the flow of information through a gating mechanism, on the other hand, strongly improves model robustness.

 %\footnote{Appendix \ref{sec:appendix2} shows results on the full data set.}

\subsection{Training on Text-Image Pairs Only}
%\label{sec:appendix2}
To compare the effect of the `average' image (see Section \ref{sec:task}) on the performance of the models, we train MM-Gate, MM-XAtt, and  MM-Gated-XAtt on tweets that are originally accompanied by an image excluding all text-only tweets; and we test on all tweets as in our original setting (text-only tweets are paired with the `average' image). The results are shown in Table \ref{tab:resultsimgsall}.
MM-Gated-XAtt is consistently the best performing model, followed by MM-Gate. However, their performance is inferior than when models are trained on all tweets using the `average' image as in the original setting. This suggests that the gate operation not only regulates the flow of information for each modality but also learns how to use the noisy modality to improve classification prediction. This result is similar to findings by \cite{arevalo2020gated}.

\section{Analysis}
%We conduct a study to better understand the properties and limitations of our multimodal MM-Gated-XAtt model for POI type prediction.
% %%%%%%%%%%%%%%%%%%%%Train With Image Only and Test All %%%%%%%%%%%%%%%%
\begin{table}[!t]
\centering
\small
\resizebox{0.38\textwidth}{!}{
\begin{tabular}{|l|l|}
\hline
\rowcolor[HTML]{C0C0C0}
\multicolumn{2}{|c|}{\cellcolor[HTML]{C0C0C0}\textbf{Text-Image Only -> All}} \\ \hline
\multicolumn{1}{|c|}{\cellcolor[HTML]{C0C0C0}\textbf{Model}} &
\multicolumn{1}{c|}{\cellcolor[HTML]{C0C0C0}\textbf{F1}} \\ \hline
%LXMERT              & 47.72 (0.98)  \\
MM-Gate             & 40.67 (0.45)
\\
MM-XAtt             & 31.00 (0.89)
\\
MM-Gated-XAtt (Ours)  &   \textbf{42.45} (2.94)   \\ \hline

\end{tabular}}
\caption{Macro F1-Score for POI type prediction. Models are trained on tweets that are originally accompanied by an image. Results are on all tweets. Best results are in bold.}
\label{tab:resultsimgsall}
\end{table}
%%%%%%%%%%%%%%%%%%%% End With Image Only %%%%%%%%%%%%%%%%%%%%%%%%%%%%%%%%%

\begin{figure}[t!]
    %\centering
    \includegraphics[width=0.48\textwidth]{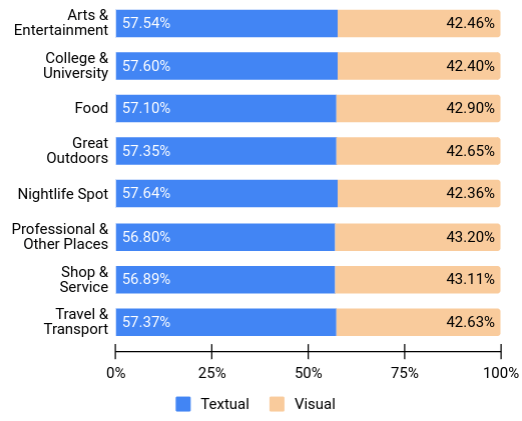}
    \caption{Average percentage of MM-Gated-XAtt activations for the textual and visual modalities for each POI category on the test set.}
    \label{fig:activations}
\end{figure}
\subsection{Modality Contribution}
To determine the influence of each modality in MM-Gated-XAtt when assigning a particular label to a tweet, we compute the average percentage of activations for the textual and visual modalities for each POI category on the test set. The outcome of this analysis is depicted in Fig. \ref{fig:activations}. As anticipated, the textual modality has a greater influence on the model prediction, which is consistent with our findings in Section \ref{sec:res}. The category where the visual modality has greater impact on the predicted label is  \emph{Professional \& Other Places} (43.20\%) followed by \emph{Shop \& Service} (43.11\%).

To examine how the visual information impacts the POI type prediction task, Fig. \ref{fig:icontrib} shows examples of posts where the contribution of the image is large while the text-only model (BERT) misclassified the POI category. We observe that the text content of Post (a) misled BERT towards \emph{Food}, probably due to the term `powder'. On the other hand, MM-Gated-XAtt can filter irrelevant information from the text, and prioritize relevant content from the image in order to assign the correct POI category for Post (a) (\emph{Great Outdoors}). Likewise, Post (b) was correctly classified by MM-Gated-XAtt as \emph{Shop \& Service} and misclassified by BERT as \emph{Arts \& Entertainment}. For this post 40\% of the contribution corresponds to the image and 60\% to text. This shows how image information can help to address the ambiguity in short texts \cite{moon-etal-2018-multimodal}, improving POI type prediction.

\subsection{Cross-attention (XAtt)} %Finally, to analyze connections between tweet text and image we visualize the XAtt in MM-Gated-XAtt.
\begin{figure}[!t]
    \footnotesize
    \centering
    %\small
    \begin{tabular}{|l|l}

        \begin{tabular}[l]{@{}l@{}} \textbf{Post (a)}\\ \#mywife finding a deep\\ first {\color{red}track} through the\\ \#powder <mention> <url>\\ \end{tabular}  &
        \begin{tabular}[l]{@{}l@{}}\textbf{Post (b)}\\ {\color{red}it's} getting cold up\\ here {\color{red}<mention>} <url>\\\\ \end{tabular}
        % \begin{tabular}[c]{@{}l@{}}\textbf{(c)} $\sim$ all of the lights $\sim$ <mention>\\ place \#pikeplacemarket <url> \end{tabular}
        \\ \\
        \cincludegraphics[scale=0.13]{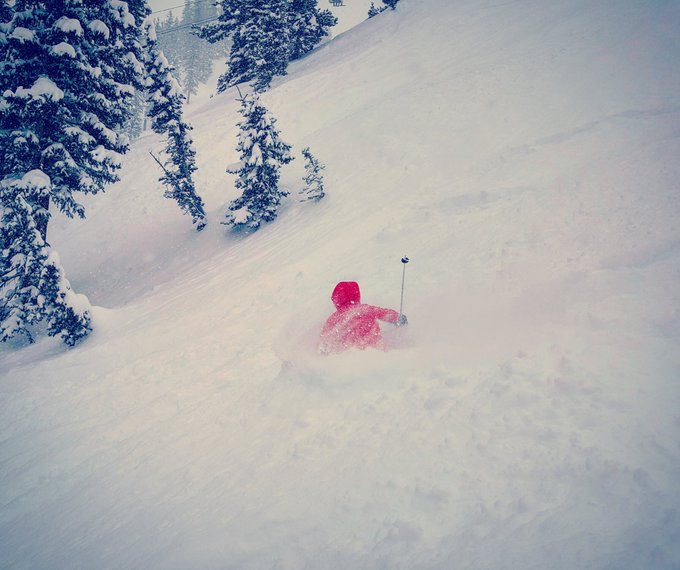}&
        %\cincludegraphics[scale=0.4]{images/analysis/shoph.png} &
        \cincludegraphics[scale=0.14]{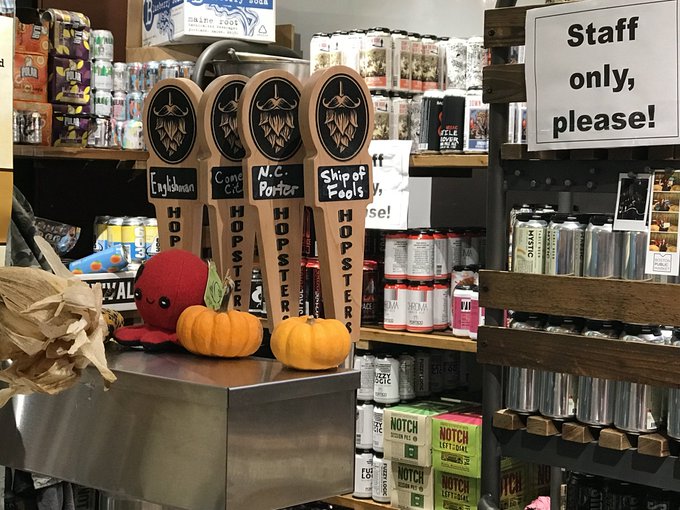} \\

        \\ \cincludegraphics[scale=0.13]{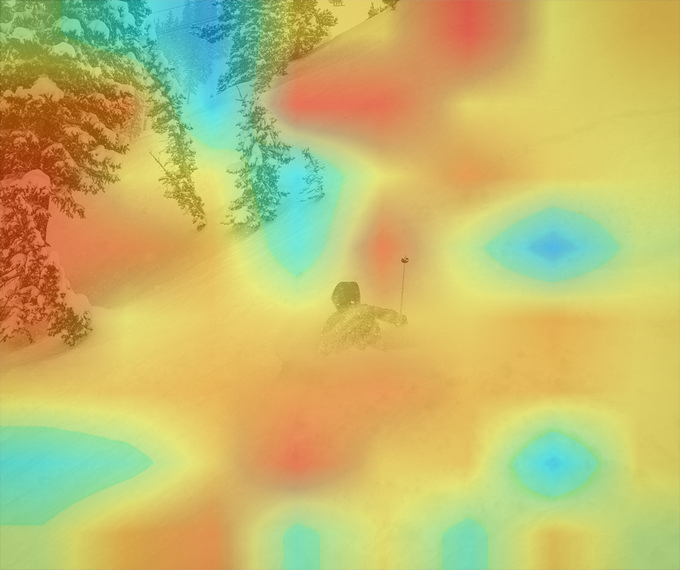} &
        \cincludegraphics[scale=0.193]{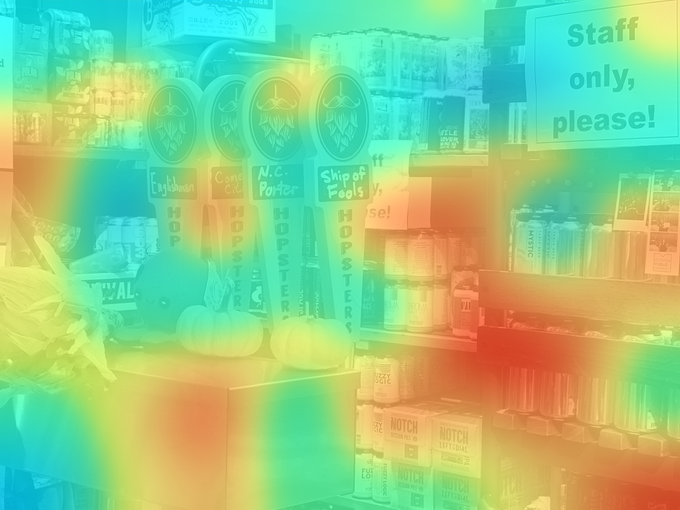}
        \\

        \begin{tabular}[l]{@{}l@{}} \\ BERT: Food \\ Ours: \textbf{Great Outdoors}\ \\ \qquad Txt: 65\% - Img: 35\% \end{tabular} &

        \begin{tabular}[l]{@{}l@{}} \\ BERT: Arts \& Entertainment\\ Ours:  \textbf{Shop \& Service} \\ \qquad Txt: 60\% - Img: 40\% \end{tabular}
        % \begin{tabular}[c]{@{}l@{}} BERT: \textbf{Shop \& Service}\\ Ours: Nightlife Spot\end{tabular}
    \end{tabular}
%    }
    \caption{POI type predictions of MM-Gated-XAtt (Ours) and BERT~\citet{sanchez-villegas-etal-2020-point} showing the contribution of each modality (\%) and the XAtt visualization. Correct predictions are in bold.}
    \label{fig:icontrib}
\end{figure}

Fig. \ref{fig:icontrib} shows examples of the XAtt visualization. We note that the model focuses on relevant nouns and pronouns (e.g. `track', `it'), which are common informative words in vision-and-language tasks \citet{tan-etal-2019-learning}. Moreover, our model focuses on relevant words such as `track' for classifying Post (a) as \emph{Great Outdoors}. Lastly, we observe that the XAtt often captures a general image information, with emphasis on specific sections for the predicted POI category such as the pine trees for \emph{Great Outdoors} and the display racks for \emph{Shop \& Service}.

%%%%%%%%%%%%%%%%%%%%% Sample Tweets %%%%%%%%%%%%%%%%%%%%%%%%%%%%%%

\renewcommand{\arraystretch}{1.2}
%%%%%%%%%%%%%%%%%%%%%%%%%%%%% END Sample Tweets %%%%%%%%%%%%%%%%%%

\subsection{Error Analysis}
To shed light on the limitations of our multimodal MM-Gated-XAtt model for predicting POI types, we performed an analysis of misclassifications. In general, we observe that the model struggles with identifying POI categories where people might perform similar activities in each of them such as \emph{Food}, \emph{Nightlife Spot}, and \emph{Shop \& Service} similar to findings by \citet{ye2011semantic}.

\begin{figure}[!t]
    %\footnotesize
    \centering
    \small
    \begin{tabular}{|l|l}
    %\begin{tabularx}{0.95\textwidth}{|X|X}
        % \begin{tabular}[l]{@{}l@{}l@{}} \textbf{Post a)} \\visit \#stuffforstudentsat\\ \#broncoscamp today for a\\ chance to win this signed\\ by 9⃣5⃣! \#beachampion <url> \end{tabular}  &
        \begin{tabular}[l]{@{}l@{}}\textbf{Post (a)}\\ miso creamed kale with\\ mushrooms <mention> \\\\\\\end{tabular}  &
        \begin{tabular}[l]{@{}l@{}}\textbf{Post (b)}\\ celebrate the fruits of\\ \#fermentation's labor at \\ \#bostonfermentationfestival!\\ next sun 10-4 <mention> \end{tabular}
        % \begin{tabular}[c]{@{}l@{}}\textbf{(c)} $\sim$ all of the lights $\sim$ <mention>\\ place \#pikeplacemarket <url> \end{tabular}
        \\
        %\cincludegraphics[scale=0.14]{images/analysis/896362160888074240_2017-08-12-13-26-07_1.jpg}&
        %\cincludegraphics[scale=0.4]{images/analysis/shoph.png} &
        \cincludegraphics[scale=0.13]{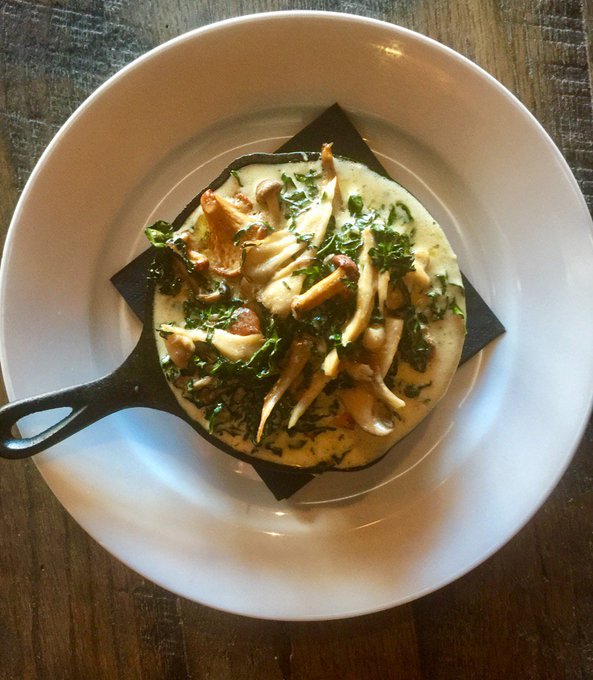} &
        \cincludegraphics[scale=0.125]{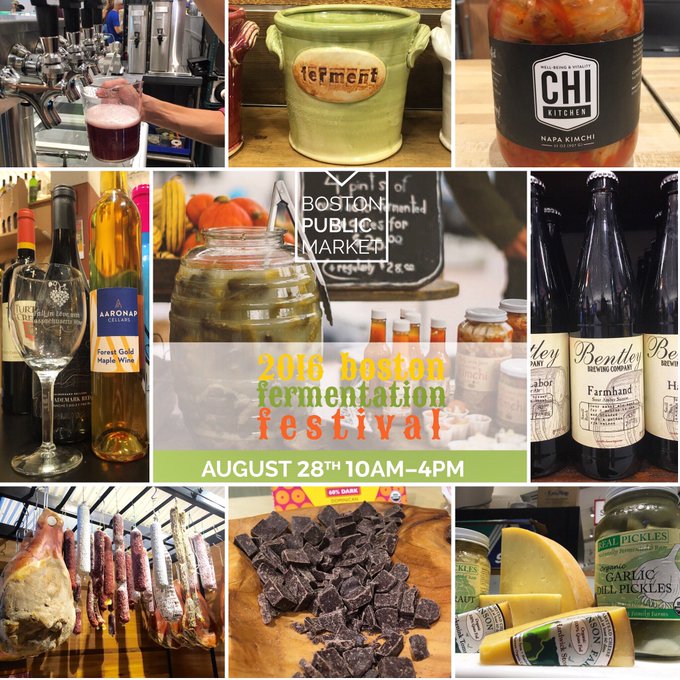}
        \\

        % \cincludegraphics[scale=0.16]{images/analysis/935778076562034688_2017-11-29-07-50-54_1.jpg}\\

        %\begin{tabular}[l]{@{}l@{}} \\ True: Great Outdoors \\ Ours: Arts \& Entertainment\end{tabular} &

        \begin{tabular}[l]{@{}l@{}} \\ True: Nightlife Spot \\ Ours:  Food \end{tabular} &
        \begin{tabular}[c]{@{}l@{}} \\ True: Shop \& Service\\ Ours: Food \end{tabular}
    \end{tabular}
%    }
    \caption{Example of misclassifications made by our MM-Gated-XAtt model. %The model struggles with identifying POI categories where people exhibit similar behaviors
    }
    \label{fig:error}
\end{figure}

Fig. \ref{fig:error} (a) and (b) show examples of tweets misclassified as \emph{Food} by the MM-Gated-XAtt model. Post (a) belongs to the category \emph{Nightlife Spot} and Post (b) belongs to the \emph{Shop \& Service} category. In both cases, the text and image content is related to the \emph{Food} category, misleading the classifier towards this POI type. Posting about food is a common practice in hospitality establishments such as restaurants and bars \cite{zhu2019post}, where customers are more likely to share content such as photos of dishes and beverages, intentionally designed to show that are associated with the particular context and lifestyle that a specific place represents \cite{homburg2015new,brunner2016impact,doi:10.1080/19368623.2020.1768195}. Similarly, Post (b) shows an example of a tweet that promotes a POI by communicating specific characteristics of the place \cite{kruk-etal-2019-integrating,aydin2020social}. To correctly classify the category of POIs, the model might need access to deeper contextual information about the locations (e.g. finer subcategories of a type of place and how POI types are related to one another).

%such as Great Outdoors and Arts \& Entertainment. Fig. \ref{fig:error} (a) is an example of a tweet that belongs to the \emph{Great Outdoors} category and was misclassified as \emph{Arts \& Entertainment} by MM-Gated-XAtt. The post promotes a POI by displaying relevant items that can be found in the place (e.g. a bag and a pigskin), a common marketing strategy of sports and entertainment venues \cite{leask2011social,wang2015sports}. In this example, the text and image content complements each other \cite{vempala-preotiuc-pietro-2019-categorizing}, and the model was misled to classify it as \emph{Arts \& Entertainment} due to the similar content to this category.

\section{Conclusion and Future Work}
This paper presents the first study on multimodal POI type classification using text and images from social media posts motivated by studies in geosemiotics, visual semiotics and cultural geography. We enrich a publicly available data set with images and we propose a multimodal model that uses: (1) a gate mechanism to control the information flow from each modality; (2) a cross-attention mechanism to align and capture the interactions between modalities. Our model achieves state-of-the-art performance for POI type prediction significantly outperforming the previous text-only model and competitive pretrained multimodal models.

In future work, we plan to perform more granular prediction of POI types and  user information to provide additional context to the models. Our models could also be used for modeling other tasks where text and images naturally occur in social media such as analyzing political ads~\cite{sanchez-villegas-etal-2021-analyzing},  parody~\cite{maronikolakis-etal-2020-analyzing} and complaints~\cite{preotiuc-pietro-etal-2019-automatically,jin-aletras-2020-complaint,jin-aletras-2021-modeling}.

%%%%%%%%%%%%%%%%%%%%% Sample Tweets %%%%%%%%%%%%%%%%%%%%%%%%%%%%%%

%\renewcommand{\arraystretch}{1.2}
%%%%%%%%%%%%%%%%%%%%%%%%%%%%% END Sample Tweets %%%%%%%%%%%%%%%%%%

\section*{Ethical Statement}
Our work complies with Twitter data policy for research,\footnote{https://developer.twitter.com/en/developer-terms/agreement-and-policy} and has received approval from the Ethics Committee of our institution (Ref. No 039665).

\section*{Acknowledgments}

We would like to thank Mali Jin, Panayiotis Karachristou and all reviewers for their valuable feedback. DSV is supported by the Centre for Doctoral Training in Speech and Language Technologies (SLT) and their Applications funded by the UK Research and Innovation grant EP/S023062/1. NA is supported by a Leverhulme Trust Research Project Grant.
\bibliography{anthology,vpoi}
\bibliographystyle{acl_natbib}

\end{document}